\documentclass[10pt,twocolumn,letterpaper]{article}

\usepackage[pagenumbers]{cvpr}
\makeatletter
\renewcommand{\paragraph}{\@startsection{paragraph}{4}{\z@}%
  {1.5bp}{-1em}{\normalfont\normalsize\bfseries}}
\makeatother

\usepackage{graphicx}
\usepackage{dblfloatfix}
\usepackage{placeins}
\usepackage{microtype}
\usepackage{amsmath}
\usepackage{amssymb}
\usepackage{amsthm}
\usepackage{booktabs}
\usepackage{eqparbox}
\newcommand{\metriccell}[2]{\eqmakebox[#1][c]{#2}}
\usepackage{multirow}
\usepackage{colortbl}
\DeclareMathVersion{table}
\SetSymbolFont{operators}{table}{OT1}{ptm}{m}{n}
\DeclareSymbolFont{uppercasegreek}{OT1}{cmr}{m}{n}
\DeclareMathSymbol{\Omega}{\mathalpha}{uppercasegreek}{"0A}
\SetMathAlphabet{\mathrm}{table}{OT1}{ptm}{m}{n}
\SetMathAlphabet{\mathbf}{table}{OT1}{ptm}{b}{n}
\SetMathAlphabet{\mathit}{table}{OT1}{ptm}{m}{it}
\AtBeginEnvironment{tabular}{\rmfamily\mathversion{table}\let\ttfamily\rmfamily}
\usepackage{algorithm}
\usepackage{listings}
\lstdefinestyle{torchpseudo}{
  language=Python,
  basicstyle=\ttfamily\footnotesize,
  keywordstyle=\bfseries,
  commentstyle=\itshape\color{black!60},
  columns=fullflexible,
  keepspaces=true,
  showstringspaces=false,
  numbers=left,
  numberstyle=\tiny\color{black!50},
  numbersep=8pt,
  xleftmargin=1.8em,
  breaklines=true,
  aboveskip=3pt,
  belowskip=0pt,
  tabsize=4
}
\usepackage{tikz}
\usetikzlibrary{arrows.meta,positioning,calc,fit,backgrounds}
\usepackage{pgfplots}
\pgfplotsset{compat=1.17}

\definecolor{cvprblue}{rgb}{0.21,0.49,0.74}
\usepackage{xr}
\usepackage[pagebackref,breaklinks,colorlinks,allcolors=cvprblue]{hyperref}

\newcommand{\secrefstyle}[1]{%
  \crefformat{#1}{\S##2##1##3}\Crefformat{#1}{\S##2##1##3}%
  \crefrangeformat{#1}{\S##3##1##4--##5##2##6}\Crefrangeformat{#1}{\S##3##1##4--##5##2##6}%
  \crefmultiformat{#1}{\S##2##1##3}{ and~##2##1##3}{, ##2##1##3}{ and~##2##1##3}%
  \Crefmultiformat{#1}{\S##2##1##3}{ and~##2##1##3}{, ##2##1##3}{ and~##2##1##3}}
\AtBeginDocument{\secrefstyle{section}\secrefstyle{subsection}\secrefstyle{subsubsection}}
\AtBeginDocument{\crefname{proposition}{Prop.}{Props.}\Crefname{proposition}{Proposition}{Propositions}}
\makeatletter
\renewcommand\paragraph{\@startsection{paragraph}{4}{\z@}{\z@}{-1em}{\normalfont\normalsize\bfseries\upshape}}
\makeatother

\newif\ifshowtodo
\showtodotrue      

\newtheorem{proposition}{Proposition}
\theoremstyle{remark}

\definecolor{gencol}{RGB}{232,125,30}
\definecolor{mechred}{RGB}{200,35,35}
\definecolor{encfill}{RGB}{207,223,245}
\definecolor{encdraw}{RGB}{95,125,175}
\definecolor{latglow}{RGB}{170,160,230}
\definecolor{panelb}{RGB}{243,247,253}
\definecolor{popA}{RGB}{70,130,180}

\newcommand{\method}{MitoWorld}

\title{\method: A Stochastic World Model of the Single-Cell Cycle\\under Genetic Perturbation}

\author{%
  Ting Long\textsuperscript{1}\quad
  Xiaojing Sun\textsuperscript{1}\quad
  Cheng Han\textsuperscript{2}\quad
  Zhicheng Ji\textsuperscript{3}\\
  Ying Nian Wu\textsuperscript{4}\quad
  Jiebo Luo\textsuperscript{5}\quad
  Heng Fan\textsuperscript{6}\quad
  Dongfang Liu\textsuperscript{1,*}\\[4pt]
  {\small
  \textsuperscript{1}Purdue University\quad
  \textsuperscript{2}University of Missouri\quad
  \textsuperscript{3}Duke University}\\
  {\small
  \textsuperscript{4}UCLA\quad
  \textsuperscript{5}University of Rochester\quad
  \textsuperscript{6}University of North Texas}\\[2pt]
  {\small\textsuperscript{*}Corresponding author}%
}

\begin{document}

\maketitle

\begin{abstract}
A cell observed after a gene knockdown has an uncertain future, both in how its appearance will evolve and when it will divide.
Existing approaches predict endpoints or population-level changes without jointly modeling individual trajectories and division timing.
In this work, we present \method, a stochastic world model that generates these future appearances as continuous evolution in a Gaussian reference latent, terminated by a division hazard and converted into images by a conditional diffusion renderer.
The ordinary transition provably preserves its Gaussian reference, tying its noise to its dynamics; paired observations identify its discrete transition where stationary marginals cannot.
The hazard turns division timing into a survival likelihood that can be tested against real cycle lengths.
On 325 held-out MitoCheck tracks spanning 17 knockdowns and a control, \method\ lowers FID by 34.6\% and the video metric JEDi by 45.6\% relative to the strongest baselines.
Its latent-space prediction intervals achieve near-nominal 90\% coverage, are 35\% narrower than those of a model with freely fitted noise.
The hazard also yields cell-cycle length distributions with KS distance close to the model-based finite-sample reference.
Testable single-cell stochastic dynamics thus offer a generative tool for time-lapse perturbation screens.
\end{abstract}

\section{Introduction}
\label{sec:intro}

The phenotype of a gene knockdown unfolds over time, and cells that start alike can follow different growth trajectories and divide at different times~\cite{elowitz2002,spencer2009,altschuler2010,sandler2015,neumann2010}.
We therefore aim to model an observed cell’s future appearance distribution under a given gene knockdown until division, that is, a world model of the single-cell cycle~\cite{bunne2024virtual,xing2026world}.
Unlike snapshot images, time-lapse screens follow individual cells over time under controlled interventions, so each generated future appearance can be scored against the real one.

\begin{figure}[!htbp]
  \centering
  \includegraphics[width=\linewidth]{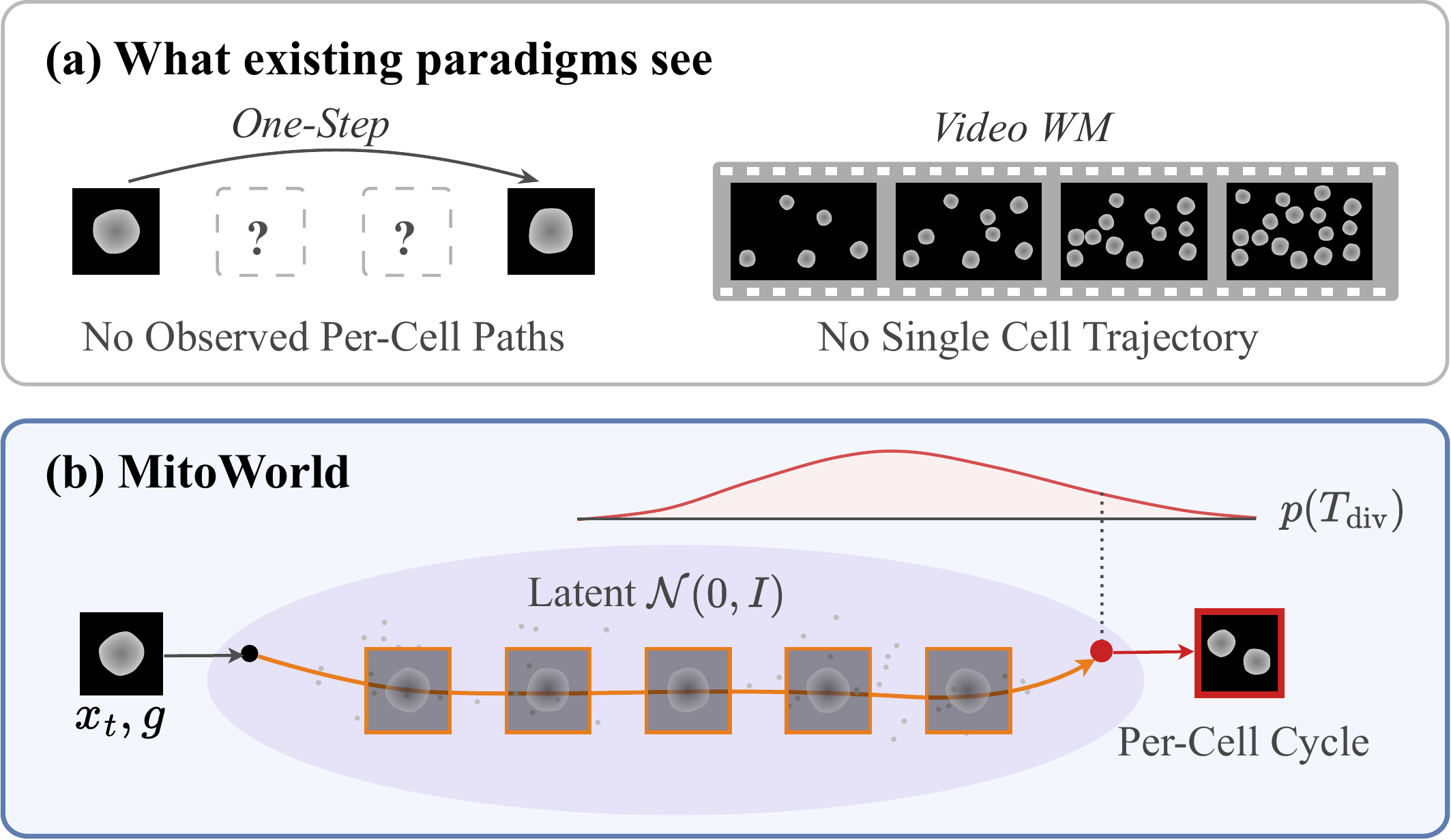}
  \caption{\textbf{Cell-cycle prediction.}
  (a) One-step endpoint predictors learn to map control populations to perturbed populations, while video world models synthesize population-level cell dynamics. Neither explicitly models how an individual cell grows or when it divides.
  (b) From one cell $x_t$ under knockdown $g$, \method\ rolls out stochastic latent trajectories, renders every step (faded box), and stops each at a division (red point) sampled from $p(T_{\mathrm{div}})$.}
  \label{fig:teaser}
\end{figure}

{
Existing approaches to predicting cellular responses largely model population-level changes, through either endpoint prediction or time-lapse video generation (\cref{fig:teaser}(a)).
First, \emph{one-step endpoint predictors} learn to map control populations to perturbed populations~\cite{lotfollahi2019,bunne2023}.
Because destructive measurements prevent repeated observations of the same cells, these methods learn from unpaired snapshots of different cell populations~\cite{weinreb2018}. Coupling populations across time can infer trajectories, but lacks directly observed same-cell correspondence~\cite{schiebinger2019,tong2020}. Methods limited to endpoint prediction additionally leave cellular growth and division timing unspecified.
Second, \emph{time-lapse video predictors} extend generative video models~\cite{ho2022video,blattmann2023,chen2024diffusion} to microscopy, synthesizing population-level cellular dynamics~\cite{holmberg2025adapting}.
Although these videos can depict mitosis, they target population-level visual dynamics rather than explicitly modeling each individual cell's growth trajectory and division time.
}

Our key insight is to make three aspects of a single-cell future explicit: cross-time dependence, predictive uncertainty, and the division event. A paired transition learns which future belongs to the same cell; a reference-constrained stochastic kernel couples the amount of noise to the learned dynamics; and a state-, age-, and gene-conditioned hazard turns division into a random stopping time. These components retain the state, transition, termination, and renderer structure of latent world models~\cite{ha2018world,hafner2023dreamerv3,scholkopf2021causal,zheng2026dsreg}, while making their predictions separately testable (\cref{fig:teaser}(b)).

We instantiate this design as \textbf{\method}. A nonlinear encoder maps frames to a Gaussian-regularized latent space, where a gene-conditioned Ornstein--Uhlenbeck transition~\cite{uhlenbeck1930} generates future states and a conditional diffusion renderer converts these states into images. The ordinary transition preserves its Gaussian reference and composes consistently across step sizes (\cref{prop:reference}). Under this reference model, paired observations identify the discrete transition kernel that population marginals cannot identify (\cref{prop:identifiable}). A hazard and two terminal transitions complete each sampled cycle with a two-daughter composite image. The method is designed to preserve individual correspondence, quantify possible futures, and predict when a trajectory terminates, rather than optimize appearance alone.

We evaluate on MitoCheck~\cite{neumann2010}, using 1,183 training and 325 test tracks spanning 17 knockdowns and a scrambled control. Our experiments evaluate image and sequence fidelity together with same-cell retrieval, then test the paired dynamics, latent uncertainty and stability, and division timing in the same order as these claims.

In summary, our contributions are threefold:
\begin{itemize}
  \item We cast the single-cell cycle under genetic perturbation as a \textbf{\textit{world-model problem}}, where a cell's state evolves stochastically and stops at a state-dependent division event, so each cell's future is a distribution over paths (\cref{sec:formulation}).
  \item We propose \method\ and prove that its transition \textbf{\textit{preserves a Gaussian reference distribution}} over any number of steps (\cref{prop:reference}) and show that its discrete transition kernel is \textbf{\textit{identifiable from per-cell pairs but not from stationary marginals}} (\cref{prop:identifiable}).
  \item On 325 held-out MitoCheck tracks, we demonstrate \method's \textbf{\textit{forecast fidelity, uncertainty quantification, and cell-cycle timing}}: FID decreases by 34.6\% relative to the strongest baseline, latent prediction intervals are 34.8\% narrower than those with freely fitted noise while maintaining near-nominal 90\% coverage, and generated cell-cycle length distributions achieve mean per-condition KS close to the model-based finite-sample reference (\cref{sec:experiments}).
\end{itemize}

\section{Related Work}
\label{sec:related}
\begin{figure*}[!tbp]
  \centering
  \newsavebox{\figTwoBox}
  \setbox\figTwoBox=\hbox{\input{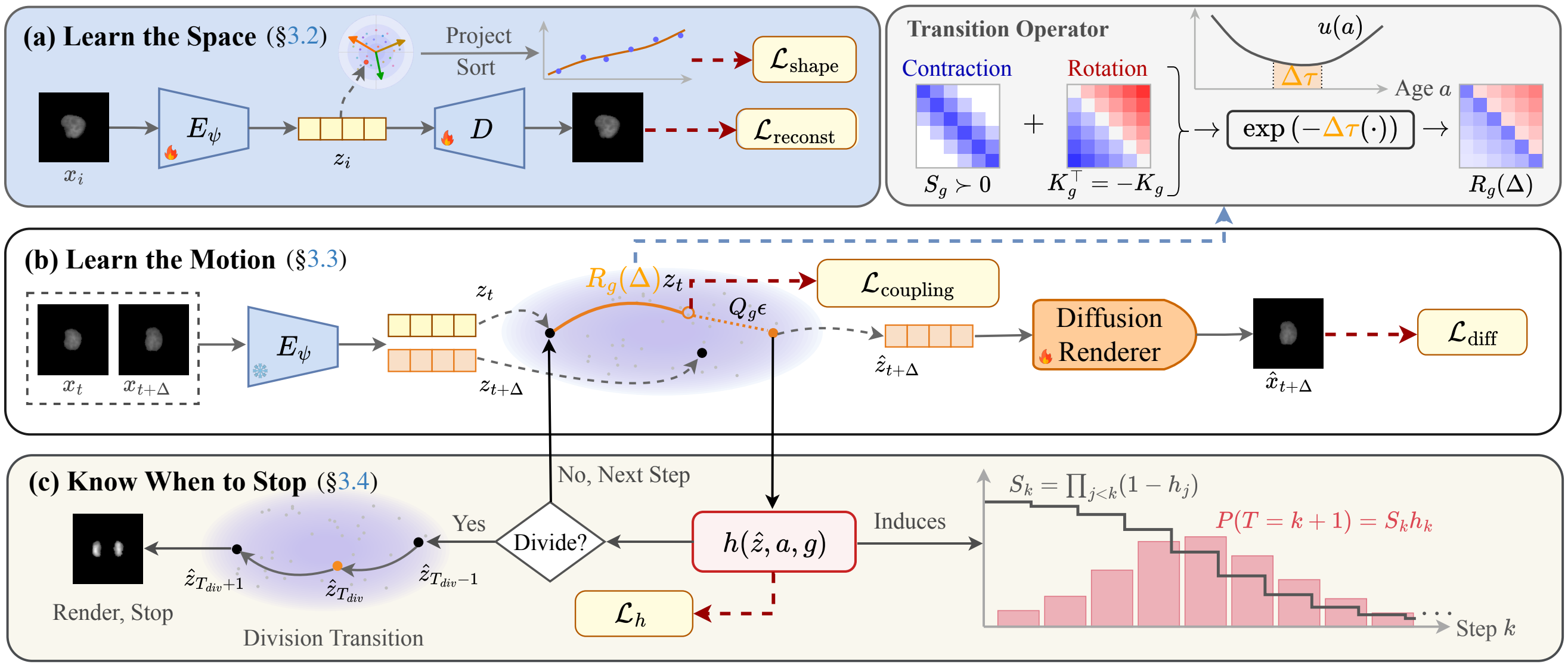}}
  \box\figTwoBox
  \caption{\textbf{The pipeline of \method.}
  (a) trains the encoder to place cell states in a Gaussian reference space (\cref{sec:space}).
  (b) freezes the encoder and moves each state with a reference-preserving transition whose operator splits into a contraction $S_g$ and a rotation $K_g$ (\cref{sec:transition}), and a diffusion renderer converts every state into an image (\cref{sec:decoder}).
  (c) uses a division hazard $h$ to determine the survival probability $S_k$ and the division-time distribution. A hazard trigger initiates the terminal transitions, which end the cycle with a two-daughter composite (\cref{sec:hazard}).}
  \label{fig:method}
\end{figure*}

\paragraph{Perturbation response prediction.}
Perturbation screens compare profiles across conditions, with transcriptional readouts of pooled screens~\cite{dixit2016perturbseq,replogle2022,yu2026pace} or morphological features from microscopy~\cite{bray2016cellpainting,caicedo2017,chandrasekaran2021}.
Predictors then map a control population to a perturbed one, in expression space~\cite{lotfollahi2019,lotfollahi2023cpa,roohani2024gears,bunne2023} or, with conditional generative models, in image space~\cite{palma2025impa,zhang2025cellflux,bourou2024phendiff,navidi2025morphodiff}, and what they pair is one population with another, because destructive profiling never observes the same cell twice~\cite{weinreb2018}.
Snapshots at successive times let optimal transport, Schr\"odinger bridges, and flow matching pair populations across time as well~\cite{schiebinger2019,tong2020,bunne2022proximal,huguet2022mioflow,koshizuka2023nlsb,tong2024sf2m}, but a coupling fitted to marginals cannot say which future belongs to which cell~\cite{lavenant2024}.
Live-cell tracking pairs each cell with its own future through the whole cycle~\cite{neumann2006,neumann2010,held2010cellcognition,zhong2012,muller2021sirna}, yet such tracks have served classification and mitotic staging rather than generation, and no model above predicts when a given cell will divide~\cite{sandler2015,cadart2018size}.
\method\ takes these same-cell pairs as supervision, which identify the transition that marginals leave ambiguous (\cref{prop:identifiable}), and models the response as a process ending at a predicted division rather than an endpoint (\cref{sec:formulation}).

\paragraph{World models for visual dynamics.}
World models learn how a system evolves from observations and simulate futures from a given state~\cite{ha2018world,hafner2019planet}, and three ingredients recur.
\emph{Reconstruction} grounds the latent in images, from masked image and video modeling to recurrent world models~\cite{he2022mae,tong2022videomae,kraus2024mae,hafner2023dreamerv3}.
\emph{Prediction} learns temporal structure through feature prediction or latent transitions~\cite{bardes2024vjepa,robine2023transformerbased,zhang2026ocstorm,peng2026eadream,balestriero2025lejepa}.
\emph{Conditioning} makes external inputs  explicit, as in action-conditioned simulators and conditional diffusion world models~\cite{bruce2024genie,yang2024unisim,valevski2025gamengen,agarwal2025cosmos,alonso2024diamond,cohen2026horizon}.
Closest to us, video diffusion has been adapted to time-lapse microscopy at the field-of-view level rather than per cell~\cite{holmberg2025adapting}.
In all of them the transition is a network, so the spread of sampled futures is a design choice rather than a tested quantity, and a rollout ends by a learned continue flag, if at all, rather than by an event with checkable timing.
\method\ keeps the state, transition, termination, and renderer of latent world models, but makes the transition an explicit reference-preserving operator (\cref{sec:transition}) and the termination a division hazard (\cref{sec:hazard}).

\paragraph{Structured stochastic dynamics and division timing.}
A cell-cycle forecast must describe how the cell evolves when it enters division. For the first problem, neural latent transitions offer flexible predictors~\cite{hafner2019planet,robine2023transformerbased,zhang2023storm}, but repeated prediction can compound errors~\cite{asadi2018lipschitz}. Learning nonlinear coordinates in which the dynamics are approximately linear makes this propagation easier to analyze~\cite{watter2015e2c,takeishi2017koopman,lusch2018koopman,otto2019lrane,azencot2020koopman}. In Gaussian-regularized coordinates~\cite{maes2026leworldmodel,lachapelle2022mechanism}, Ornstein--Uhlenbeck transition attenuates existing deviations, and preserving the Gaussian reference ties its noise covariance to the mean operator~\cite{uhlenbeck1930,sarkka2019sde,jang2026latentprojection,orvieto2023lru}. \method\ learns a gene-conditioned transition of this form from same-cell trajectories and evaluates per-cell prediction error  over long rollouts (\cref{sec:exp_calibration}).
Division ends the trajectory, making its timing a time-to-event problem. In discrete-time survival analysis, the hazard is the probability of an event at a given step conditional on its not having occurred earlier~\cite{cox1972,singer1993}. Drawing on this framework~\cite{sandler2015,cadart2018size}, \method\ treats division as a random stopping time with a hazard conditioned on the generated state, cell age, and gene.  We compare the predicted division-time distributions with observed cycle lengths (\cref{sec:exp_division}).

\section{Method}
\label{sec:method}
\method\ is a world model of the single-cell cycle built from a latent state space, a stochastic transition, a division hazard, and a diffusion renderer, which \crefrange{sec:space}{sec:decoder} describe in turn and \cref{fig:method} summarizes.
We first state the problem (\cref{sec:formulation}) and close with training and inference (\cref{sec:training}).

\subsection{Problem formulation}
\label{sec:formulation}

Let $x_{0:T}$ denote a single-cell track under gene-knockdown condition $g$, with $a_k$ denoting the cell's age at frame $k$. Given an observed frame $x_t$, its age $a_t$, and condition $g$, we jointly sample future appearances and division timing:
\begin{equation}
\big(\hat x_{t+1:T_{\mathrm{div}}+1},T_{\mathrm{div}}\big)
\sim p_\Theta(\,\cdot\mid x_t,a_t,g).
\label{eq:factor}
\end{equation}
As summarized in \cref{fig:method}, an encoder initializes the latent state $z_t=E_\psi(x_t)$, and a gene-conditioned stochastic transition propagates it through time. A state-, age-, and gene-conditioned hazard stochastically triggers two terminal transitions, producing chromatin condensation at $T_{\mathrm{div}}$ and a two-daughter composite frame at $T_{\mathrm{div}}+1$. A conditional diffusion renderer maps the resulting states to images. This construction models the joint evolution of the same cell across time rather than independent population marginals.

\subsection{Learn the latent space}
\label{sec:space}
We learn a nonlinear representation with a Gaussian reference, aiming to capture complex image-level variation in the encoder while modeling cellular evolution through linear stochastic dynamics. An encoder $E_\psi$ maps each frame $x$ to a latent state $z = E_\psi(x) \in \mathbb{R}^d$, and a decoder $D$ reconstructs the frame using a weighted L1 loss:
\begin{equation}
\mathcal{L}_{\text{reconst}} = \langle w_p|\hat x_p-x_p|\rangle_p.
\label{eq:reconst}
\end{equation}
Here $x_p$ and $\hat x_p$ are the normalized target and reconstructed intensities at pixel $p$, $w_p$ is weight parameter to focus more on foreground pixels, and the brackets average over pixels and the training batch.

We regularize the pooled latent distribution toward $\mathcal{N}(0,I)$ using:
\begin{equation}
\mathcal{L}_{\text{shape}}
=\mathcal{L}_{\text{SW}}
+\frac{1}{d}\sum_{j=1}^{d}
\bigl(|\hat\mu_j|+|\hat\sigma_j-1|\bigr),
\label{eq:shape}
\end{equation}
where $\mathcal{L}_{\text{SW}}$ is the sliced Wasserstein discrepancy to $\mathcal{N}(0,I)$~\cite{kolouri2019swae,tolstikhin2018wae}, and $\hat\mu_j,\hat\sigma_j$ are the batch mean and standard deviation of the $j$-th latent coordinate. The constraint applies to pooled states across conditions, without requiring each gene-specific distribution to be Gaussian. We evaluate its contribution to long-horizon forecasting in \cref{sec:exp_rollout}.

\subsection{Learn how cells move}
\label{sec:transition}

Predicting an individual cell requires learning how its current state constrains its future, beyond matching population appearance. We therefore learn stochastic dynamics from same-cell frame pairs. In the learned representation, we use a linear conditional mean to propagate the state and tie its uncertainty to the dynamics through the Gaussian reference. This design combines directional evolution with contraction, while allowing progression speed to vary with cell age.

\paragraph{Reference-preserving stochastic transition.}
Before the division hazard triggers, we propagate the latent state using a gene-conditioned Ornstein--Uhlenbeck (OU) transition~\cite{uhlenbeck1930}:
\begin{equation}
\begin{aligned}
\hat z_{k+\Delta}
&=R_g(\Delta)z_k+Q_g(\Delta)\epsilon_k,
\quad \epsilon_k\sim\mathcal{N}(0,I),\\
R_g(\Delta)
&=e^{-\Delta A_g},
\quad Q_g(\Delta)=\bigl(I-R_gR_g^\top\bigr)^{1/2}.
\end{aligned}
\label{eq:transition}
\end{equation}

The conditional mean advances the current state, while fresh noise generates alternative futures. Requiring the transition to preserve $\mathcal{N}(0,I)$ determines its covariance from $R_g$, linking uncertainty to the decay of initial-state information rather than fitting an independent covariance. 

\paragraph{Structured operator.}
To allow directional changes in the latent mean during cell-cycle progression, we combine a contractive component with a rotational component:
\begin{equation}
\begin{gathered}
A_g = S_g + K_g,\quad S_g \succ 0,\quad K_g^\top = -K_g,\\
S_g = \Lambda + U\,\mathrm{diag}(\beta_g)\,U^\top,
\quad K_g = U\,\Omega_g\,U^\top,
\end{gathered}
\label{eq:operator}
\end{equation}
where $\Lambda\succ0$ is a shared diagonal matrix and $U\in\mathbb{R}^{d\times r}$ has orthonormal columns with $r\ll d$. In the deterministic mean flow, $S_g$ controls norm decay, while the skew-symmetric $K_g$ changes direction without directly changing the state norm. The rotational component therefore allows temporal evolution beyond contraction alone. To share statistical strength across perturbations, gene-dependent coefficients $\beta_g\geq0$ and $\Omega_g=-\Omega_g^\top$ modulate these components within a common low-rank subspace, rather than fitting a separate full-dimensional operator for each condition.

\cref{prop:reference} formalizes how the structured operator and tied covariance preserve the Gaussian reference and ensure consistent composition across step sizes.

\begin{proposition}[Reference preservation]
\label{prop:reference}
For a fixed gene $g$, $S_g\succ0$ makes $R_g(\Delta)$ a strict contraction for every $\Delta>0$, so $Q_g(\Delta)$ is well defined. The transition preserves $\mathcal{N}(0,I)$, and consecutive steps of lengths $\Delta_1,\Delta_2$ equal one step of length $\Delta_1+\Delta_2$ in distribution.
\end{proposition}

\paragraph{Biological clock.}
To capture age-dependent progression rates, we use a learned monotone biological clock \(\tau\) to convert chronological age into effective transition time:
\begin{equation}
\Delta_k=\tau(a_{k+1})-\tau(a_k).
\label{eq:clock}
\end{equation}

The clock adjusts evolution speed across cycle ages while retaining the same gene-conditioned operator. Its average rate over a reference cycle is normalized to one.  Because clock increments add, the composition property also
holds for age-dependent effective steps. Reference preservation
is exact for the ordinary kernel, while the encoded distribution
only approximately matches the Gaussian reference.

\paragraph{Learning cross-time correspondence.}
Reference preservation alone does not specify how a particular cell evolves: different transitions can share the same time-point marginals. We therefore fit the operator and clock using frames $(x_u,x_v)$ from the same cell cycle with the encoder frozen, learning the cross-time dependence needed for individual forecasting:
\begin{equation}
\mathcal{L}_{\text{coupling}}
=\mathbb{E}\!\left[
\frac{1}{d}
\left\|R_g(\Delta)z_u-z_v\right\|_2^2
\right],
\label{eq:coupling}
\end{equation}
where $z_u=E_\psi(x_u)$, $z_v=E_\psi(x_v)$, and $\Delta=\tau(a_v)-\tau(a_u)$. Under the reference model, \cref{prop:identifiable} establishes
that these pairs identify the discrete transition kernel.

\begin{proposition}[Identifiability from pairs]
\label{prop:identifiable}
For a fixed gene $g$, let $z_t\sim\mathcal{N}(0,I)$ evolve
under \cref{eq:transition}. The marginal of $z_{t+k\Delta}$
is $\mathcal{N}(0,I)$ for every $k$ and every admissible
operator $A_g$, so time-point marginals cannot identify
the transition. Paired states identify $R_g(\Delta)$, since
$\mathbb{E}[z_{t+\Delta}z_t^\top]=R_g(\Delta)$.
This determines the full conditional
$p(z_{t+\Delta}\mid z_t)
=\mathcal{N}\bigl(R_g(\Delta)z_t,\,
I-R_g(\Delta)R_g(\Delta)^\top\bigr)$
and hence the joint distribution over paths on the
corresponding time grid.
\end{proposition}

Thus, reference preservation constrains repeated stochastic evolution, while paired supervision identifies cross-time dependence. Recovering the continuous generator $A_g$ requires additional spectral conditions. We evaluate paired supervision and operator structure in \cref{sec:exp_ablation}, and long-horizon behavior in \cref{sec:exp_rollout}.

\subsection{Know when to stop}
\label{sec:hazard}
The ordinary OU transition models gradual cellular evolution over extended horizons, while completing a cell cycle requires predicting when division begins and capturing its abrupt morphological changes. We therefore combine a conditioned hazard for event timing with two terminal transitions for condensation and two-daughter formation.

\paragraph{Predicting division timing.}
Cell age signals cycle progression, while the state and gene condition modulate division propensity. We express the probability of entering condensation at the next frame through a discrete-time hazard~\cite{singer1993,cox1972}:
\begin{equation}
h(z,a,g)=\sigma\!\left(
b+\sum_{j=0}^{K_{\text{age}}-1}\gamma_j\varphi_j(a)
+w_z^\top z+w_g^\top e_g
\right),
\label{eq:hazard}
\end{equation}
where $\sigma$ is the sigmoid function, $\{\varphi_j(a)\}_{j=0}^{K_{\text{age}}-1}$ are the $K_{\text{age}}$ Gaussian age basis functions, $e_g$ is the gene embedding shared with the transition, and $b,\gamma_j,w_z,w_g$ are learned parameters. This formulation allows cells of the same age to have different division probabilities according to their state and perturbation condition.

Let $h_j=h(z_j,a_j,g)$ along an ordinary OU path starting from an observation at frame $t$. As illustrated in \cref{fig:method}(c), survival through frame $k\geq t$ requires no trigger at frames $t$ through $k-1$, giving $S_k=\prod_{j=t}^{k-1}(1-h_j)$ with $S_t=1$. A first trigger at $k$ produces condensation at $T_{\mathrm{div}}=k+1$ with pathwise probability $S_kh_k$. Averaging over ordinary OU paths conditional on $(z_t,a_t,g)$, denoted by $\mathbb{E}_t$, gives:
\begin{equation}
\begin{aligned}
P(T_{\mathrm{div}}>k\mid z_t,a_t,g) &= \mathbb{E}_t[S_k],\\
P(T_{\mathrm{div}}=k+1\mid z_t,a_t,g) &= \mathbb{E}_t[S_kh_k].
\end{aligned}
\label{eq:survival}
\end{equation}

For each set of $N$ labeled hazard decisions, we minimize binary cross-entropy:
\begin{equation}
\mathcal{L}_h = -\frac{1}{N}\sum_{i=1}^{N}
\bigl[y_i\log h_i+(1-y_i)\log(1-h_i)\bigr],
\label{eq:hazard_loss}
\end{equation}
where $h_i$ is the predicted hazard, and $y_i=1$ at the frame immediately before condensation and $0$ at earlier frames. We average the losses on observed states and training rollout states with equal weights. At sampling time, the first hazard trigger switches the rollout from OU dynamics to the division transitions described below.

\paragraph{The division transition.}
To represent abrupt division-associated changes separately from ordinary evolution, we fit two affine Gaussian transitions:
\begin{equation}
p_\eta^{(r)}(z'\mid z)
=\mathcal{N}(z';J_rz+b_r,V_r),
\quad r\in\{M,C\},
\label{eq:division_transition}
\end{equation}
where $J_r$, $b_r$, and $V_r$ are the stage-specific linear map, offset, and conditional covariance. Stage $M$ maps the pre-condensation state to condensed chromatin, and stage $C$ maps that state to the two-daughter composite. Their parameters $\eta=\{J_r,b_r,V_r\}_{r\in\{M,C\}}$ are fitted from training terminal triples. A hazard trigger at frame $k$ produces condensation at $k+1=T_{\mathrm{div}}$ and the composite at $k+2$, after which the rollout stops. Together, the hazard and terminal transitions specify when ordinary evolution ends and how the sampled cycle completes. Division timing and conditioning-factor ablations are evaluated in \cref{sec:exp_division}.

\subsection{What cells look like}
\label{sec:decoder}
A conditional diffusion renderer~\cite{ho2020ddpm,peebles2023dit,preechakul2022diffae} converts the latent states generated by the ordinary and terminal transitions into images. The diffusion renderer uses $x$-prediction~\cite{li2025backtobasics} to directly predict the clean frame $\hat x_\phi(x_s,s,z,a,g,c)$ from the noisy target $x_s$ at diffusion step $s$, the latent state $z$, cell age $a$, gene condition $g$, and source image $c$. We train it with a weighted MSE:
\begin{equation}
\mathcal{L}_{\text{diff}} = \mathbb{E}\Big[
w_{\mathrm{time}}(s)\tfrac{1}{HW}\textstyle\sum_{p=1}^{HW}
w_{\mathrm{focal},p}(\hat x_{\phi,p}-x_p)^2\Big],
\label{eq:diff}
\end{equation}
where $H$ and $W$ are the image dimensions, $w_{\mathrm{time}}$ emphasizes low-noise steps, and $w_{\mathrm{focal},p}$ emphasizes pixels with larger reconstruction errors. We evaluate rendering in \cref{tab:table1_main_comparison} and illustrate sampled trajectories in \cref{fig:rollouts}.

\subsection{Learning the space, then the motion}
\label{sec:training}
We first learn the latent representation through reconstruction and Gaussian regularization. We then freeze the encoder and learn the motion and rendering models in fixed latent coordinates. The objectives for these two stages are:
\begin{equation}
\begin{aligned}
  \mathcal{L}_{\text{space}} &= \mathcal{L}_{\text{reconst}} + \lambda_{\text{shape}}\mathcal{L}_{\text{shape}}, \\
  \mathcal{L}_{\text{motion}} &= \mathcal{L}_{\text{coupling}} + \lambda_{\text{diff}}\mathcal{L}_{\text{diff}}.
\end{aligned}
\label{eq:objectives}
\end{equation}
Here $\lambda_{\text{shape}}$ and $\lambda_{\text{diff}}$ represent the weight of $\mathcal{L}_{\text{shape}}$ and $\mathcal{L}_{\text{diff}}$, respectively. We fit the coupling independently of the hazard using latent pairs encoded from observed frames of the same cell.  Once the coupling is trained, we keep the encoder and coupling fixed and fit the division hazard with $\mathcal{L}_h$ using both observed and rollout states (\cref{sec:hazard}). Delaying hazard fitting avoids exposing it to unreliable predictions from early coupling training, while rollout states expose it to the model-generated inputs it will encounter at inference.

At inference, we initialize the latent state as $z_t=E_\psi(x_t)$. At each step, we sample a trigger from the hazard. If no event is sampled, the ordinary transition (\cref{eq:transition}) advances the state. Upon a trigger, the two terminal transitions (\cref{eq:division_transition}) generate the final states in sequence. The diffusion renderer converts the sampled states into images, and the rollout ends after the composite frame. Repeating this procedure from the same observation yields joint samples of future image trajectories and division times for the same cell.

\begin{table*}[!t]
\caption{\textbf{Image prediction and biological features} on 325 test MitoCheck~\cite{neumann2010} tracks. FID measures frame fidelity at +3h, +10h, pre-division (\texttt{div}), and across future frames (\texttt{all}); R@1 measures cell-specific retrieval, and JEDi measures sequence realism. The five biological categories average marginal W1 distances. \textcolor{red}{BEST} and \textcolor{blue}{SECOND} results are highlighted.}
\label{tab:table1_main_comparison}
\centering\small
\setlength{\tabcolsep}{1.5pt}
\begin{tabular*}{\textwidth}{@{\extracolsep{\fill}}cl||cccccc|ccccc@{}}
\hline
\rowcolor{gray!20}
 & Method & \metriccell{TableOneMetric}{{\scriptsize FID +3h$\downarrow$}} & \metriccell{TableOneMetric}{{\scriptsize FID +10h$\downarrow$}} & \metriccell{TableOneMetric}{{\scriptsize FID div$\downarrow$}} & \metriccell{TableOneMetric}{{\scriptsize FID all$\downarrow$}} & \metriccell{TableOneMetric}{{\scriptsize R@1$\uparrow$}} & \metriccell{TableOneMetric}{{\scriptsize JEDi$\downarrow$}} & \metriccell{TableOneMetric}{{\scriptsize Size$\downarrow$}} & \metriccell{TableOneMetric}{{\scriptsize Intensity$\downarrow$}} & \metriccell{TableOneDist}{{\scriptsize Intensity dist.$\downarrow$}} & \metriccell{TableOneMetric}{{\scriptsize Texture$\downarrow$}} & \metriccell{TableOneMetric}{{\scriptsize Shape$\downarrow$}} \\
\hline\hline
\multirow{4}{*}{\rotatebox[origin=c]{90}{\scriptsize Generative}} & CellFlux {\scriptsize\textcolor{gray!55}{\textnormal{[ICML25]}}}~\cite{zhang2025cellflux} & \metriccell{TableOneMetric}{74.72} & \metriccell{TableOneMetric}{99.39} & \metriccell{TableOneMetric}{97.96} & \metriccell{TableOneMetric}{85.74} & \metriccell{TableOneMetric}{\textcolor{blue}{37.33\%}} & \metriccell{TableOneMetric}{7.14} & \metriccell{TableOneMetric}{0.3056} & \metriccell{TableOneMetric}{\textcolor{blue}{0.0789}} & \metriccell{TableOneDist}{0.4073} & \metriccell{TableOneMetric}{0.1343} & \metriccell{TableOneMetric}{0.1476} \\
 & IMPA {\scriptsize\textcolor{gray!55}{\textnormal{[Nat.~Commun.25]}}}~\cite{palma2025impa} & \metriccell{TableOneMetric}{68.61} & \metriccell{TableOneMetric}{\textcolor{blue}{64.65}} & \metriccell{TableOneMetric}{\textcolor{blue}{55.10}} & \metriccell{TableOneMetric}{54.38} & \metriccell{TableOneMetric}{5.67\%} & \metriccell{TableOneMetric}{6.50} & \metriccell{TableOneMetric}{\textcolor{blue}{0.1626}} & \metriccell{TableOneMetric}{4.0374} & \metriccell{TableOneDist}{0.5267} & \metriccell{TableOneMetric}{2.4320} & \metriccell{TableOneMetric}{0.1812} \\
 & PhenDiff {\scriptsize\textcolor{gray!55}{\textnormal{[MICCAI24]}}}~\cite{bourou2024phendiff} & \metriccell{TableOneMetric}{306.09} & \metriccell{TableOneMetric}{300.45} & \metriccell{TableOneMetric}{279.16} & \metriccell{TableOneMetric}{289.75} & \metriccell{TableOneMetric}{3.59\%} & \metriccell{TableOneMetric}{9.61} & \metriccell{TableOneMetric}{1.3789} & \metriccell{TableOneMetric}{3.6650} & \metriccell{TableOneDist}{0.8561} & \metriccell{TableOneMetric}{3.2474} & \metriccell{TableOneMetric}{4.7492} \\
 & MorphoDiff {\scriptsize\textcolor{gray!55}{\textnormal{[ICLR25]}}}~\cite{navidi2025morphodiff} & \metriccell{TableOneMetric}{104.21} & \metriccell{TableOneMetric}{123.30} & \metriccell{TableOneMetric}{110.95} & \metriccell{TableOneMetric}{112.68} & \metriccell{TableOneMetric}{6.46\%} & \metriccell{TableOneMetric}{9.78} & \metriccell{TableOneMetric}{0.5144} & \metriccell{TableOneMetric}{3.8396} & \metriccell{TableOneDist}{\textcolor{blue}{0.2490}} & \metriccell{TableOneMetric}{2.9829} & \metriccell{TableOneMetric}{\textcolor{blue}{0.1356}} \\
\hline
\multirow{6}{*}{\rotatebox[origin=c]{90}{\scriptsize World Model}} & DIAMOND {\scriptsize\textcolor{gray!55}{\textnormal{[NeurIPS24]}}}~\cite{alonso2024diamond} & \metriccell{TableOneMetric}{89.36} & \metriccell{TableOneMetric}{95.69} & \metriccell{TableOneMetric}{73.22} & \metriccell{TableOneMetric}{80.09} & \metriccell{TableOneMetric}{5.64\%} & \metriccell{TableOneMetric}{8.31} & \metriccell{TableOneMetric}{0.3126} & \metriccell{TableOneMetric}{3.4623} & \metriccell{TableOneDist}{0.7152} & \metriccell{TableOneMetric}{3.7944} & \metriccell{TableOneMetric}{0.3780} \\
 & DreamerV3 {\scriptsize\textcolor{gray!55}{\textnormal{[Nature25]}}}~\cite{hafner2023dreamerv3} & \metriccell{TableOneMetric}{176.59} & \metriccell{TableOneMetric}{183.83} & \metriccell{TableOneMetric}{176.49} & \metriccell{TableOneMetric}{177.08} & \metriccell{TableOneMetric}{18.92\%} & \metriccell{TableOneMetric}{\textcolor{blue}{3.75}} & \metriccell{TableOneMetric}{0.2734} & \metriccell{TableOneMetric}{0.0789} & \metriccell{TableOneDist}{0.7617} & \metriccell{TableOneMetric}{0.1191} & \metriccell{TableOneMetric}{0.4845} \\
 & TWM {\scriptsize\textcolor{gray!55}{\textnormal{[ICLR23]}}}~\cite{robine2023transformerbased} & \metriccell{TableOneMetric}{163.91} & \metriccell{TableOneMetric}{182.36} & \metriccell{TableOneMetric}{185.30} & \metriccell{TableOneMetric}{174.80} & \metriccell{TableOneMetric}{14.15\%} & \metriccell{TableOneMetric}{5.00} & \metriccell{TableOneMetric}{0.3918} & \metriccell{TableOneMetric}{0.1472} & \metriccell{TableOneDist}{0.7166} & \metriccell{TableOneMetric}{0.1567} & \metriccell{TableOneMetric}{0.6853} \\
 & OC-STORM {\scriptsize\textcolor{gray!55}{\textnormal{[ICLR26]}}}~\cite{zhang2026ocstorm} & \metriccell{TableOneMetric}{237.07} & \metriccell{TableOneMetric}{250.05} & \metriccell{TableOneMetric}{244.82} & \metriccell{TableOneMetric}{242.06} & \metriccell{TableOneMetric}{18.54\%} & \metriccell{TableOneMetric}{6.54} & \metriccell{TableOneMetric}{0.3164} & \metriccell{TableOneMetric}{0.1870} & \metriccell{TableOneDist}{0.4785} & \metriccell{TableOneMetric}{\textcolor{blue}{0.1000}} & \metriccell{TableOneMetric}{0.4861} \\
 & EADream {\scriptsize\textcolor{gray!55}{\textnormal{[ICLR26]}}}~\cite{peng2026eadream} & \metriccell{TableOneMetric}{190.13} & \metriccell{TableOneMetric}{201.46} & \metriccell{TableOneMetric}{205.69} & \metriccell{TableOneMetric}{196.87} & \metriccell{TableOneMetric}{8.59\%} & \metriccell{TableOneMetric}{8.87} & \metriccell{TableOneMetric}{0.1998} & \metriccell{TableOneMetric}{0.1653} & \metriccell{TableOneDist}{0.6971} & \metriccell{TableOneMetric}{0.2068} & \metriccell{TableOneMetric}{0.5444} \\
 & HI {\scriptsize\textcolor{gray!55}{\textnormal{[ICLR26]}}}~\cite{cohen2026horizon} & \metriccell{TableOneMetric}{\textcolor{blue}{56.81}} & \metriccell{TableOneMetric}{66.31} & \metriccell{TableOneMetric}{57.37} & \metriccell{TableOneMetric}{\textcolor{blue}{54.36}} & \metriccell{TableOneMetric}{36.72\%} & \metriccell{TableOneMetric}{4.73} & \metriccell{TableOneMetric}{0.2662} & \metriccell{TableOneMetric}{0.0988} & \metriccell{TableOneDist}{0.2503} & \metriccell{TableOneMetric}{0.1091} & \metriccell{TableOneMetric}{0.1582} \\
\hline
 & \textbf{\method} & \metriccell{TableOneMetric}{\textcolor{red}{\textbf{20.71}}} & \metriccell{TableOneMetric}{\textcolor{red}{\textbf{46.30}}} & \metriccell{TableOneMetric}{\textcolor{red}{\textbf{44.01}}} & \metriccell{TableOneMetric}{\textcolor{red}{\textbf{35.54}}} & \metriccell{TableOneMetric}{\textcolor{red}{\textbf{41.87\%}}} & \metriccell{TableOneMetric}{\textcolor{red}{\textbf{2.04}}} & \metriccell{TableOneMetric}{\textcolor{red}{\textbf{0.1349}}} & \metriccell{TableOneMetric}{\textcolor{red}{\textbf{0.0612}}} & \metriccell{TableOneDist}{\textcolor{red}{\textbf{0.1586}}} & \metriccell{TableOneMetric}{\textcolor{red}{\textbf{0.0591}}} & \metriccell{TableOneMetric}{\textcolor{red}{\textbf{0.1021}}} \\
\hline
\end{tabular*}
\end{table*}

\section{Experiments}
\label{sec:experiments}
We first describe the experimental setup(\cref{sec:setup}) and then compare \method\ with strong baselines on single-cell forecasting performance and further complement this evaluation with qualitative analyses in image and latent space (\cref{sec:main_results}). 
After that, we ablate paired supervision and operator structure, assess calibration and rollout stability, and test the contribution of hazard conditioning to division timing in \crefrange{sec:ablation_analysis}{sec:exp_division} and close with failure case analysis (\cref{sec:failure_cases}).

\begingroup
\let\experimenttopic\paragraph
\renewcommand{\subsubsection}[1]{\experimenttopic{#1.}}
\makeatletter
\renewcommand\paragraph{\@startsection{paragraph}{4}{\z@}{\z@}{-1em}{\normalfont\normalsize\bfseries\upshape}}
\makeatother

\subsection{Experimental setup}
\label{sec:setup}
\paragraph{Dataset.}
We use a subset of MitoCheck~\cite{neumann2010}, a genome-wide time-lapse RNAi screen of HeLa cells expressing H2B-GFP and imaged every 30 minutes. Our subset comprises 17 gene knockdown conditions and one negative control. After excluding censored tracks, we retain 1,508 complete single-cell cycles with recorded birth and division events, partitioned at the track level into 1,183 training and 325 test trajectories. For each tracked cell, we rotate the image to align the segmentation mask’s major axis vertically and set pixels outside the mask to zero. We center the mask’s bounding-box crop on a ($96\times96$) grayscale canvas, discarding the oversized crops.

\paragraph{Baselines.}
We compare four perturbation-conditioned generative models: CellFlux~\cite{zhang2025cellflux}, IMPA~\cite{palma2025impa}, PhenDiff~\cite{bourou2024phendiff}, and MorphoDiff~\cite{navidi2025morphodiff} and further compare against six competitive world models: DIAMOND~\cite{alonso2024diamond}, DreamerV3~\cite{hafner2023dreamerv3}, TWM~\cite{robine2023transformerbased}, OC-STORM~\cite{zhang2026ocstorm}, EADream~\cite{peng2026eadream}, and HI~\cite{cohen2026horizon}, which capture temporal dynamics for multi-step visual prediction. We adapt their predictive components to single-cell image forecasting.

\paragraph{Reproducibility.}
We train \method\ for 100 epochs on a single NVIDIA H200 GPU using AdamW with a learning rate of $10^{-4}$ and a batch size of 128. We generate images using source-conditioned DDIM with 16 sampling steps.

\subsection{Predicting individual cellular futures}
\label{sec:main_results}
\label{sec:exp_percell}

\paragraph{Appearance and individual correspondence.}
We first assess whether the model predicts individual cellular futures and improves image fidelity and sequence-level temporal realism while preserving correspondence to the same cell. In \cref{tab:table1_main_comparison}, \method\ achieves an overall FID of 35.54 and JEDi of 2.04, improving on the strongest respective baselines (54.36 and 3.75) by 34.6\% and 45.6\%. Image-feature retrieval reaches 41.87\%, compared with 37.33\% for the strongest baseline. This is consistent with explicit stochastic dynamics constraining evolution across time and same-cell paired supervision learning the cross-time dependence needed for individual forecasting (\cref{sec:transition,prop:identifiable}). In addition, for interpretable cellular-feature evaluation, we follow the 17-feature taxonomy of LIVECell~\cite{edlund2021livecell}: \emph{size} (area, perimeter, Feret's diameter), \emph{intensity} (mean, minimum, maximum, fifth and 95th percentiles), \emph{intensity distribution} (skewness, kurtosis, normalized weighted centroid), \emph{texture} (intensity standard deviation), and \emph{shape} (eccentricity, roundness, circularity, solidity, aspect ratio). \method\ achieves the lowest category-averaged marginal W1 distance in all five biological categories. Together, these results show that \method\ not only generates temporally evolving cellular trajectories but also captures the morphological feature distributions of future cells.


\begin{figure*}[!t]
  \centering
  \input{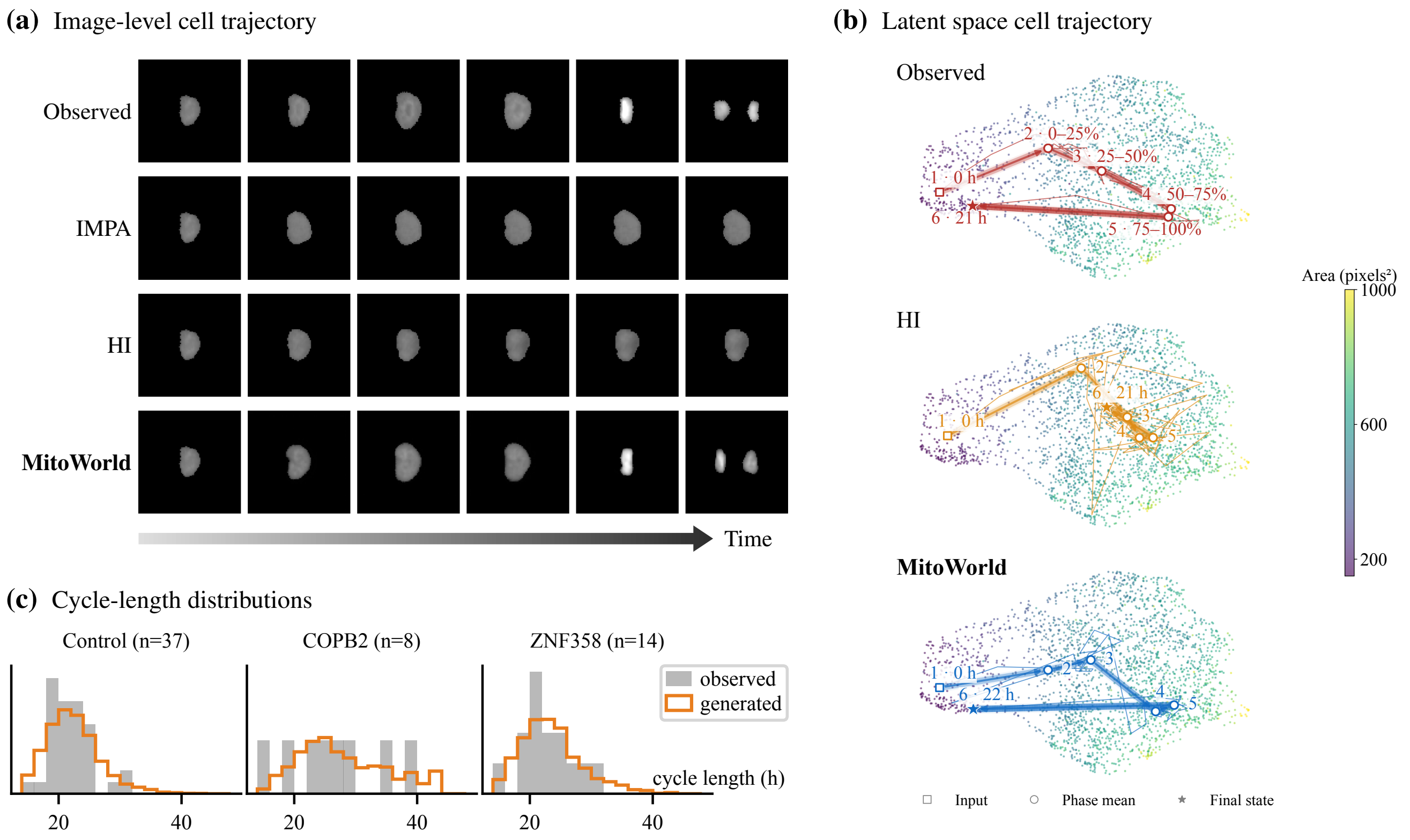}
  \caption{\textbf{Single-cell growth, division, and latent trajectories.}
  \textbf{(a) Image-level cell trajectory.} Observed trajectory and predictions from strong baselines IMPA and HI, and our \method.
  \textbf{(b) Latent space cell trajectory.} Top to bottom: the observed encoded trajectory, HI native-latent trajectory, and \method\ coupling rollout, shown on the same frozen UMAP. Thin lines show full state trajectories; thick arrows connect phase-averaged states. Background points are colored by cell area (pixels$^2$), reflecting variation in cell size. \textbf{(c) Cycle-length distributions.} Observed and generated cycle lengths for three conditions.}
  \label{fig:rollouts}
\end{figure*}

\paragraph{Image- and latent-level trajectories.}
In \cref{fig:rollouts}a, we compare observed snapshots with predictions against two strong baselines from \cref{tab:table1_main_comparison}, IMPA and HI, at corresponding stages of the cell cycle. IMPA captures size changes but rounds the cell excessively, whereas HI yields smoother shape transitions with little of the observed growth. \method\ captures changes in both size and shape, consistent with its lowest distances on both metrics in \cref{tab:table1_main_comparison}, and continues through chromatin condensation to two-daughter formation.
Panel (b) visualizes state evolution on a fixed UMAP. Before division, the observed and \method\ trajectories progress through similar regions, while HI follows a broader, less ordered path through the central region. During the hazard-triggered terminal transitions, \method\ reproduces the observed late return toward regions of smaller nuclear area, consistent with chromatin condensation before division. These examples complement \cref{fig:rollouts}a by illustrating how the model organizes individual futures across both gradual progression and abrupt division events. \Cref{fig:rollouts}c shows that \method\ captures gene-dependent differences in cell-cycle length distributions. Together, these results further support \method's ability to forecast both how a cell evolves and when it divides.

\paragraph{Human expert evaluation.}
Experts compared anonymized sequences from \method\ and strong baselines from \cref{tab:table1_main_comparison} on five criteria (\cref{tab:table2_expert_evaluation}): validity (plausible cell appearance), temporal coherence (natural changes across frames), agreement with observations (morphological match at corresponding stages), division morphology (plausible terminal division), and overall quality (visual quality of the full sequence).
Although HI achieved the highest validity score for realistic cell appearance, its low scores for agreement and temporal coherence suggesting limited fidelity to the observed cells’ morphological changes over time. \method\ scored above 4 on all five criteria and led on four, suggesting that it preserves realistic cell appearance while generating biologically consistent growth and division trajectories.

\begin{table}[!htbp]
\captionsetup{font=footnotesize}
\caption{\textbf{Human expert evaluation} on a five-point Likert scale (1: strongly disagree; 5: strongly agree).}
\label{tab:table2_expert_evaluation}
\centering\footnotesize
\setlength{\tabcolsep}{1.65pt}
\begin{tabular}{l||ccccc}
\hline
\rowcolor{gray!20}
Model & \metriccell{L}{Valid.$\uparrow$} & \metriccell{L}{Temp.$\uparrow$} & \metriccell{L}{Agr.$\uparrow$} & \metriccell{L}{Div.$\uparrow$} & \metriccell{L}{Overall$\uparrow$} \\
\hline\hline
IMPA & \metriccell{L}{2.27\textsubscript{\textcolor{black!50}{$\pm0.73$}}} & \metriccell{L}{2.63\textsubscript{\textcolor{black!50}{$\pm0.69$}}} & \metriccell{L}{1.87\textsubscript{\textcolor{black!50}{$\pm0.57$}}} & \metriccell{L}{1.03\textsubscript{\textcolor{black!50}{$\pm0.08$}}} & \metriccell{L}{2.67\textsubscript{\textcolor{black!50}{$\pm0.72$}}} \\
CellFlux & \metriccell{L}{3.57\textsubscript{\textcolor{black!50}{$\pm0.45$}}} & \metriccell{L}{2.93\textsubscript{\textcolor{black!50}{$\pm1.02$}}} & \metriccell{L}{2.20\textsubscript{\textcolor{black!50}{$\pm0.79$}}} & \metriccell{L}{1.13\textsubscript{\textcolor{black!50}{$\pm0.18$}}} & \metriccell{L}{3.50\textsubscript{\textcolor{black!50}{$\pm0.74$}}} \\
DIAMOND & \metriccell{L}{2.47\textsubscript{\textcolor{black!50}{$\pm0.19$}}} & \metriccell{L}{1.30\textsubscript{\textcolor{black!50}{$\pm0.32$}}} & \metriccell{L}{1.60\textsubscript{\textcolor{black!50}{$\pm0.25$}}} & \metriccell{L}{1.03\textsubscript{\textcolor{black!50}{$\pm0.08$}}} & \metriccell{L}{1.60\textsubscript{\textcolor{black!50}{$\pm0.37$}}} \\
OC-STORM & \metriccell{L}{3.77\textsubscript{\textcolor{black!50}{$\pm0.68$}}} & \metriccell{L}{1.80\textsubscript{\textcolor{black!50}{$\pm0.46$}}} & \metriccell{L}{2.63\textsubscript{\textcolor{black!50}{$\pm0.64$}}} & \metriccell{L}{1.00\textsubscript{\textcolor{black!50}{$\pm0.00$}}} & \metriccell{L}{2.10\textsubscript{\textcolor{black!50}{$\pm0.09$}}} \\
HI & \metriccell{L}{\textbf{4.30}\textsubscript{\textcolor{black!50}{$\pm0.51$}}} & \metriccell{L}{2.43\textsubscript{\textcolor{black!50}{$\pm0.63$}}} & \metriccell{L}{2.00\textsubscript{\textcolor{black!50}{$\pm0.72$}}} & \metriccell{L}{1.03\textsubscript{\textcolor{black!50}{$\pm0.08$}}} & \metriccell{L}{2.97\textsubscript{\textcolor{black!50}{$\pm0.94$}}} \\
\hline
\textbf{\method} & \metriccell{L}{4.03\textsubscript{\textcolor{black!50}{$\pm0.36$}}} & \metriccell{L}{\textbf{4.23}\textsubscript{\textcolor{black!50}{$\pm0.36$}}} & \metriccell{L}{\textbf{4.93}\textsubscript{\textcolor{black!50}{$\pm0.15$}}} & \metriccell{L}{\textbf{4.93}\textsubscript{\textcolor{black!50}{$\pm0.09$}}} & \metriccell{L}{\textbf{4.33}\textsubscript{\textcolor{black!50}{$\pm0.26$}}} \\
\hline
\end{tabular}

\end{table}

\subsection{Why paired and structured dynamics?}

\label{sec:ablation_analysis}
\label{sec:exp_ablation}
To examine the roles of paired supervision and transition design, we compare the full operator (\cref{eq:operator}) with shuffled-pair supervision, scalar ($\rho I$) and diagonal operators, removal of the rotational component $K_g$, removal of gene conditioning, and a uniform biological clock (\cref{tab:table5_coupling_ablation}).

\paragraph{Biological readouts from observed states.}
We first test whether the encoded state $z$ carries phenotypic information about an individual cell beyond what is predictable from its age and gene condition. We fit linear ridge probes on the training tracks using either the frozen encoder states of observed frames or age and gene as inputs.
We weight frames inversely by track length so that each track contributes equally. On the 325 held-out tracks, the latent readout achieves higher test $R^2$ for area, eccentricity, mean intensity, and texture contrast than the age-and-gene baseline. 
\begin{table}[H]
\caption{\textbf{Phenotypic information in the frozen encoder.} Equal-track test $R^2$ for linear readouts from age and gene or from $z$. Brackets give 95\% lineage-root bootstrap intervals.}
\label{tab:table4_state_readout}
\centering\footnotesize
\setlength{\tabcolsep}{1.9pt}
\begin{tabular}{l||cc}
\hline
\rowcolor{gray!20}
Feature & \metriccell{I}{Age + gene} & \metriccell{I}{Latent $z$} \\
\hline\hline
Area & \metriccell{I}{$0.483$} & \metriccell{I}{$0.804$} \\
 & \metriccell{I}{\scriptsize $[0.425,0.542]$} & \metriccell{I}{\scriptsize $[0.773,0.832]$} \\
Eccentricity & \metriccell{I}{$0.050$} & \metriccell{I}{$0.802$} \\
 & \metriccell{I}{\scriptsize $[0.019,0.074]$} & \metriccell{I}{\scriptsize $[0.779,0.822]$} \\
Mean intensity & \metriccell{I}{$0.270$} & \metriccell{I}{$0.885$} \\
 & \metriccell{I}{\scriptsize $[0.208,0.326]$} & \metriccell{I}{\scriptsize $[0.869,0.897]$} \\
Texture contrast & \metriccell{I}{$0.268$} & \metriccell{I}{$0.693$} \\
 & \metriccell{I}{\scriptsize $[0.224,0.306]$} & \metriccell{I}{\scriptsize $[0.653,0.724]$} \\
\hline
\end{tabular}

\end{table}
These readouts ((\cref{tab:table4_state_readout})) measure phenotypic information in observed states and complement the forecast correspondence evaluated by retrieval in \cref{sec:exp_percell}. The higher readout scores indicate that $z$ encodes cell-specific phenotypic variation beyond what age and gene alone predict, motivating the ablations below that test whether paired supervision and structured dynamics preserve correspondence between successive states of the same cell.

\paragraph{Learning cross-time correspondence.}
We test whether paired supervision and structured dynamics preserve cross-time dependence for individual forecasting. In \cref{tab:table5_coupling_ablation}, shuffling targets within gene/age strata preserves endpoint populations but reduces latent retrieval from 41.23\% to 31.18\%. This supports same-cell pairing: cross-time dependence identifies the discrete transition under the reference model, whereas population marginals cannot (\cref{prop:identifiable}). Shuffled pairs nevertheless achieve lower NLL, showing that likelihood can improve even as individual forecasts worsen. A broader predictive covariance can reduce covariance-normalized residuals despite weaker correspondence.

\begin{table}[!htbp]
\caption{\textbf{Effects of coupling components on latent-state prediction}.  Red and blue mark the \textcolor{red}{BEST} and \textcolor{blue}{SECOND BEST} values.}
\label{tab:table5_coupling_ablation}
\centering\small
\setlength{\tabcolsep}{1.9pt}
\begin{tabular}{l||ccc}
\hline
\rowcolor{gray!20}
Variant & \metriccell{O}{Err$\downarrow$} & \metriccell{O}{NLL$\downarrow$} & \metriccell{O}{R@1$_z$$\uparrow$} \\
\hline\hline
$\rho I$ & \metriccell{O}{0.9093} & \metriccell{O}{1.3760} & \metriccell{O}{36.62\%} \\
Diagonal & \metriccell{O}{0.9072} & \metriccell{O}{1.3740} & \metriccell{O}{36.24\%} \\
w/o $K_g$ & \metriccell{O}{0.8996} & \metriccell{O}{1.3575} & \metriccell{O}{39.66\%} \\
\hline
Shuffled pairs & \metriccell{O}{0.9185} & \metriccell{O}{\textcolor{red}{1.3343}} & \metriccell{O}{31.18\%} \\
\hline
No gene cond. & \metriccell{O}{\textcolor{blue}{0.8932}} & \metriccell{O}{1.3596} & \metriccell{O}{40.55\%} \\
\hline
Uniform clock & \metriccell{O}{0.9002} & \metriccell{O}{1.4229} & \metriccell{O}{\textcolor{blue}{41.16\%}} \\
\hline
\textbf{Full} & \metriccell{O}{\textcolor{red}{0.8903}} & \metriccell{O}{\textcolor{blue}{1.3439}} & \metriccell{O}{\textcolor{red}{41.23\%}} \\
\hline
\end{tabular}
\end{table}

\paragraph{Evaluating the structured operator.}
Scalar and diagonal dynamics reach 36.62\% and 36.24\% retrieval, respectively, below Full's 41.23\%, with higher prediction errors. These restricted operators cannot express the same coupled evolution of latent coordinates. Removing $K_g$ retains symmetric coupling but reduces retrieval to 39.66\% and increases error to 0.8996. This additional contrast supports the value of rotation beyond contraction: $S_g$ controls norm decay, while the skew term $K_g$ changes direction without directly increasing the state norm (\cref{eq:operator,eq:transition}). The results are thus consistent with the structured transition expressing directional progression while retaining contractive dynamics, complementing the qualitative trajectories in \cref{fig:rollouts}b.

\paragraph{Adjusting condition and progression speed.}
Removing gene conditioning degrades performance across all three metrics, supporting its role in adapting the shared dynamics to different perturbations. A uniform clock leaves retrieval nearly unchanged but raises NLL to 1.4229, supporting clock modulation’s role in improving state and distribution prediction by adjusting effective evolution speed across cycle ages (\cref{eq:clock}). Together, these comparisons support complementary roles for paired supervision, structured motion, and condition- and age-dependent modulation.

\subsection{Are stochastic futures calibrated and stable?}
\label{sec:exp_calibration}

\paragraph{Uncertainty from dynamics.}
With the mean transition fixed, we compare no noise, isotropic noise, a freely fitted covariance, and the reference-tied covariance of \method\ (\cref{tab:table6_noise_calibration}). 
\begin{table}[!htbp]
\caption{\textbf{Calibration results of latent futures.} }
\label{tab:table6_noise_calibration}
\centering\small
\setlength{\tabcolsep}{1.9pt}
\begin{tabular}{lc||cccc}
\hline
\rowcolor{gray!20}
Noise & \metriccell{P}{Horizon} & \metriccell{P}{\texttt{CRPS}$\downarrow$} & \metriccell{P}{\texttt{ES}$\downarrow$} & \metriccell{P}{\texttt{Cov90}} & \metriccell{P}{\texttt{Width}$\downarrow$} \\
\hline\hline
None & \metriccell{P}{\multirow{4}{*}{\texttt{+3h}}} & \metriccell{P}{$0.62$} & \metriccell{P}{$6.07$} & \metriccell{P}{--} & \metriccell{P}{--} \\
Isotropic & \metriccell{P}{} & \metriccell{P}{$0.48$} & \metriccell{P}{$4.72$} & \metriccell{P}{$98.02\%$} & \metriccell{P}{$4.03$} \\
Free & \metriccell{P}{} & \metriccell{P}{$0.47$} & \metriccell{P}{$4.69$} & \metriccell{P}{$97.88\%$} & \metriccell{P}{$3.96$} \\
\textbf{\method} & \metriccell{P}{} & \metriccell{P}{$\mathbf{0.44}$} & \metriccell{P}{$\mathbf{4.36}$} & \metriccell{P}{$\mathbf{89.72\%}$} & \metriccell{P}{$\mathbf{2.58}$} \\
\hline
None & \metriccell{P}{\multirow{4}{*}{\texttt{div}}} & \metriccell{P}{$0.79$} & \metriccell{P}{$7.55$} & \metriccell{P}{--} & \metriccell{P}{--} \\
Isotropic & \metriccell{P}{} & \metriccell{P}{$0.59$} & \metriccell{P}{$5.84$} & \metriccell{P}{$99.28\%$} & \metriccell{P}{$5.15$} \\
Free & \metriccell{P}{} & \metriccell{P}{$0.58$} & \metriccell{P}{$5.78$} & \metriccell{P}{$99.21\%$} & \metriccell{P}{$5.01$} \\
\textbf{\method} & \metriccell{P}{} & \metriccell{P}{$\mathbf{0.54}$} & \metriccell{P}{$\mathbf{5.35}$} & \metriccell{P}{$\mathbf{88.92\%}$} & \metriccell{P}{$\mathbf{3.05}$} \\
\hline
\end{tabular}
\end{table}

By setting $\Sigma_g=I-R_gR_g^\top$, \method\ links injected uncertainty to the loss of initial-state information, constraining uncertainty estimation without separately fitting a covariance matrix. At +3h and the pre-division horizon, coordinatewise coverage is 89.72\% and 88.92\%, with intervals about 35\% and 39\% narrower than Free, together with lower CRPS and ES~\cite{gneiting2007scoring}. These results support our dynamics-constrained uncertainty model, which yields sharper predictions with coverage close to the nominal 90\% level (\cref{eq:transition}). Overcoverage by Free and Isotropic indicates that wider intervals alone are not more informative.

\paragraph{Stable long-horizon dynamics.}
\label{sec:exp_rollout}
\cref{fig:stability} compares our OU transition with latent autoregression (AR) and the same OU structure fitted without Gaussian encoder regularization. \method\ maintains closer agreement with observed future distributions over longer horizons, while its per-cell error levels off rather than growing rapidly as in latent AR. Removing $\mathcal{L}_{\text{shape}}$ yields lower short-horizon per-cell error but greater long-horizon drift. These trends support combining a Gaussian-regularized representation with OU dynamics: the linear conditional mean propagates latent states, while contraction and the tied covariance control moment growth (\cref{eq:transition,prop:reference}). Thus, linear stochastic dynamics in the learned latent space can sustain stable long-horizon evolution without a nonlinear autoregressive transition.

\begin{figure}[!htbp]
 \centering
 \includegraphics[width=\linewidth]{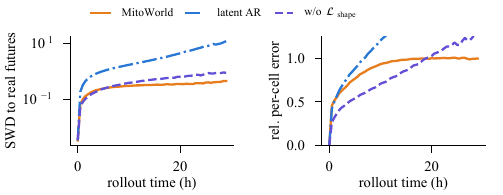}
 \caption{\textbf{Long-horizon latent dynamics}. }
 \label{fig:stability}
\end{figure}

\subsection{Predicting division events}
\label{sec:exp_division}
Having evaluated cellular state evolution, we now examine division timing. Building on the cycle-length distributions in \cref{fig:rollouts}c, we assess how state, age, and gene conditioning contribute to the hazard model (\cref{eq:hazard}).
The full hazard yields an average KS distance of 0.2229 across the 18 conditions. Even samples drawn from the same distribution can differ when only a few cells are available per condition. To estimate the size of this difference, we drew 200 simulated test sets from the model's predicted cycle lengths, matching the actual number of test cells in each condition. The resulting average KS distance was 0.2234, close to the 0.2229 measured on the real test set.
We then ablate each conditioning factor using Brier score, per-condition KS, and terminal FID (\cref{tab:table3_division_timing}). Removing age increases KS from 0.2229 to 0.4743, identifying age as the main timing signal. Removing state or gene conditioning also worsens all three metrics, supporting their complementary roles in adjusting age-dependent division propensity. The full hazard also yields the lowest terminal FID, indicating closer agreement in terminal appearance distributions.

\begin{table}[!htbp]
\caption{\textbf{Effects of the factors on division timing prediction.}}
\label{tab:table3_division_timing}
\centering\footnotesize
\setlength{\tabcolsep}{1.9pt}
\begin{tabular}{ccc||ccc}
\hline
\rowcolor{gray!20}
\metriccell{M}{\mbox{State $z$}} & \metriccell{M}{\mbox{Age $a_k$}} & \metriccell{M}{\mbox{Gene $e_g$}} & \metriccell{M}{\texttt{Brier}$\downarrow$} & \metriccell{M}{\texttt{KS}$\downarrow$} & \metriccell{M}{\texttt{FID}$\downarrow$} \\
\hline\hline
\metriccell{M}{} & \metriccell{M}{$\checkmark$} & \metriccell{M}{$\checkmark$} & \metriccell{M}{0.0211} & \metriccell{M}{0.2263} & \metriccell{M}{30.61} \\
\metriccell{M}{$\checkmark$} & \metriccell{M}{} & \metriccell{M}{$\checkmark$} & \metriccell{M}{0.0220} & \metriccell{M}{0.4743} & \metriccell{M}{32.95} \\
\metriccell{M}{$\checkmark$} & \metriccell{M}{$\checkmark$} & \metriccell{M}{} & \metriccell{M}{0.0202} & \metriccell{M}{0.2414} & \metriccell{M}{30.65} \\
\hline
\metriccell{M}{$\checkmark$} & \metriccell{M}{$\checkmark$} & \metriccell{M}{$\checkmark$} & \metriccell{M}{\textbf{0.0201}} & \metriccell{M}{\textbf{0.2229}} & \metriccell{M}{\textbf{30.50}} \\
\hline
\end{tabular}

\end{table}

 Together with the preceding experiments, these results support a unified model of individual cellular evolution, dynamics-constrained uncertainty, and division timing over a complete cell cycle.

\subsection{Failure case analysis}
\label{sec:failure_cases}

\method\ achieves the lowest discrepancy between observed and generated size distributions in \cref{tab:table1_main_comparison}, however, on closer inspection, we found that \method\ can underestimate nuclear growth in cells with unusually large nuclei. In the held-out example in \cref{fig:failure_case}, the mean observed area is $1.73$ times the mean across test tracks for the same gene. 
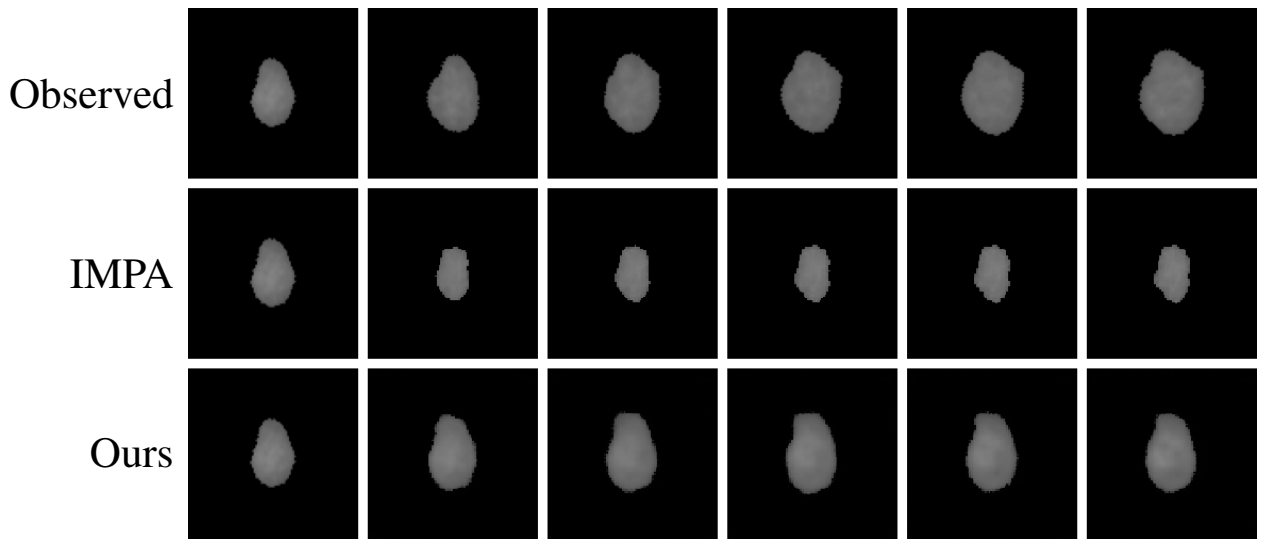
\begin{figure}[!htbp]
  \centering
  \pgfmathsetlengthmacro{\failureStripWidth}{\linewidth-36pt}
  \pgfmathsetlengthmacro{\failureRowPitch}{\failureStripWidth*0.168159204}
  \begin{tikzpicture}[failurelabel/.style={anchor=east,inner sep=0pt,font=\fontsize{8.1167}{9.7400}\selectfont}]
    \node[anchor=west,inner sep=0pt] at (0,0)
      {\includegraphics[width=\failureStripWidth]{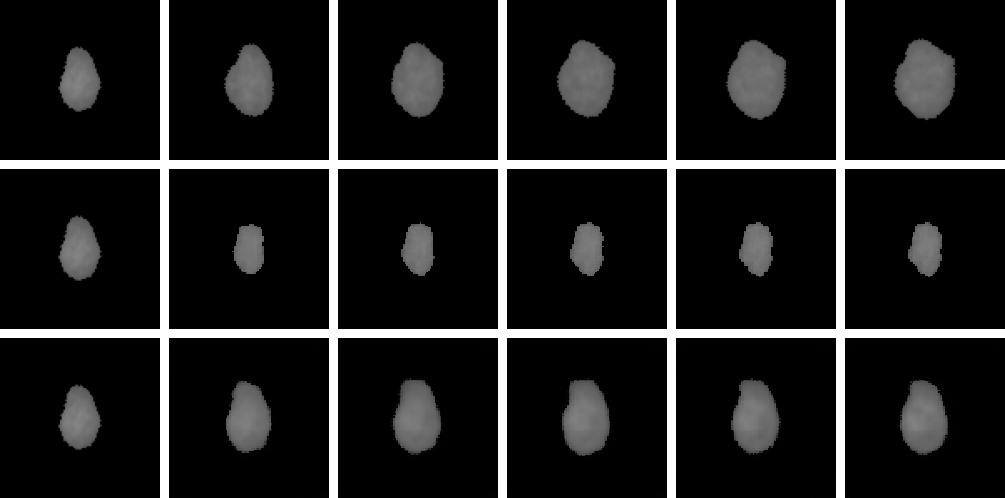}};
    \node[failurelabel] at (-2.3094pt,\failureRowPitch) {Observed};
    \node[failurelabel] at (-2.3094pt,0) {IMPA};
    \node[failurelabel] at (-2.3094pt,-\failureRowPitch) {Ours};
  \end{tikzpicture}
  \caption{\textbf{Underestimated nuclear growth.} Both methods generate trajectories from the first observed frame.}
  \label{fig:failure_case}
\end{figure}
From the shared observation, \method\ preserves the overall nuclear shape and predicts the observed widening, but underestimates its extent. This case also challenges IMPA, a strong baseline on the size metric, which predicts initial shrinkage while the observed nucleus widens. For \method, this underestimation may reflect the influence of more common growth patterns on the learned dynamics. Improving generalization to rare single-cell growth trajectories is a direction for future work.

\FloatBarrier
\endgroup
\section{Conclusion}
\label{sec:conclusion}
\method\ encodes an observed cell in a Gaussian-regularized latent space and learns an Ornstein--Uhlenbeck transition from same-cell pairs. This linear-Gaussian transition supports stable long-horizon rollouts, a division hazard determines when the cell enters division, and a diffusion renderer maps predicted states to microscopy images. On held-out MitoCheck tracks, \method\ reduces overall FID by 34.6\% and JEDi by 45.6\% relative to the strongest respective baselines. We learn cell dynamics from paired observations and evaluate on known knockdowns in one dataset and cell line. Future work includes improving generalization both in rare single-cell case and unseen gene conditions, and adapting the dynamics to unpaired snapshots.

{\small
\bibliographystyle{ieeenat_fullname}
\bibliography{main}
}

\end{document}